\documentclass[10pt,twocolumn,llncs]{article}

\usepackage{iccv}
\usepackage{times}
\usepackage{epsfig}
\usepackage{graphicx}
\usepackage{amsmath}
\usepackage{amssymb}

\usepackage{multirow}
\usepackage{booktabs}
\usepackage{graphicx}
\usepackage{amssymb}

\usepackage[hyphens]{url}
\usepackage{times}
\usepackage{epsfig}
\usepackage{graphicx}
\usepackage{amsmath}
\usepackage{float}
\usepackage{booktabs}
\usepackage{array, blkarray, makecell}
\usepackage{tabularx}
\usepackage{amsfonts}
\usepackage{float}

\usepackage{color}
\newcommand{\cB}[1]{{\color{black}{#1}}}
\newcommand{\cG}[1]{{\color{black}{#1}}}
\newcommand{\cR}[1]{{\color{black}{#1}}}

\usepackage[pagebackref=true,breaklinks=true,letterpaper=true,colorlinks,bookmarks=false]{hyperref}

\usepackage{hyperref} 
\makeatletter 
\newcommand{\printfnsymbol}[1]{%
\textsuperscript{\@fnsymbol{#1}}%
} 
\makeatother

\iccvfinalcopy 

\def\iccvPaperID{****} 

\ificcvfinal\fi

\begin{document}

\title{Describing and Localizing Multiple Changes with Transformers}

\author{Yue Qiu$^1$\thanks{equal contribution},\ Shintaro Yamamoto$^{1,2}$\printfnsymbol{1},\ Kodai Nakashima$^1$,\ Ryota Suzuki$^1$, \\
\ Kenji Iwata$^1$,\ Hirokatsu Kataoka$^1$, \ Yutaka Satoh$^1$ \\
$^1$National Institute of Advanced Industrial Science and Technology (AIST), $^2$Waseda University\\
{\tt\small \{qiu.yue, yamamoto.shintaro, nakashima.kodai, ryota.suzuki,}\\
{\tt\small kenji.iwata, hirokatsu.kataoka, yu.satou\}@aist.go.jp}

}

\maketitle
\ificcvfinal\thispagestyle{empty}\fi

\begin{abstract}

Change captioning tasks aim to detect changes in image pairs observed before and after a scene change and generate a natural language description of the changes. Existing change captioning studies have mainly focused on a single change.However, detecting and describing multiple changed parts in image pairs is essential for enhancing adaptability to complex scenarios. We solve the above issues from three aspects: (i) We propose a \cB{simulation}-based multi-change captioning dataset; (ii) We benchmark existing state-of-the-art methods of single change captioning on multi-change captioning; (iii) We further propose Multi-Change Captioning transformers (MCCFormers) that identify change regions by densely correlating different regions in image pairs and dynamically determines the related change regions with words in sentences. The proposed method obtained the highest scores on four conventional change captioning evaluation metrics for multi-change captioning. Additionally, our proposed method can separate attention maps for each change and performs well with respect to change localization. Moreover, the proposed framework outperformed the previous state-of-the-art methods on an existing change captioning benchmark, CLEVR-Change, by a large margin (+6.1 on BLEU-4 and +9.7 on CIDEr scores), indicating its general ability in change captioning tasks. The code and dataset are available at the project page \footnote{ \url{https://cvpaperchallenge.github.io/Describing-and-Localizing-Multiple-Change-with-Transformers}}.

\end{abstract}

\section{Introduction}

Detecting and describing the changed parts in scenes at different times is essential in various scenarios, such as urbanization analysis
\cite{zhang2002urban,yang2003urban,hegazy2015monitoring}, resource management \cite{coppin1996digital,kennedy2009remote,khan2017forest,saha2019unsupervised,daudt2018fully}, updating street-view maps for navigation \cite{alcantarilla2018street,sakurada2013detecting}, damage detection \cite{sakurada2015change,fujita2017damage}, video surveillance \cite{jhamtani2018learning}, and robotic applications \cite{herbst2011toward,ambrucs2014meta}. Recently, Jhamtani and Berg-Kirkpatrick \cite{jhamtani2018learning} proposed the change captioning task to describe changes from image pairs of before and after scene changes. 
\cB{Describing change is useful for extracting semantic contents and conveying information to humans.} 

\begin{figure}
\centering
\includegraphics[width=\linewidth]{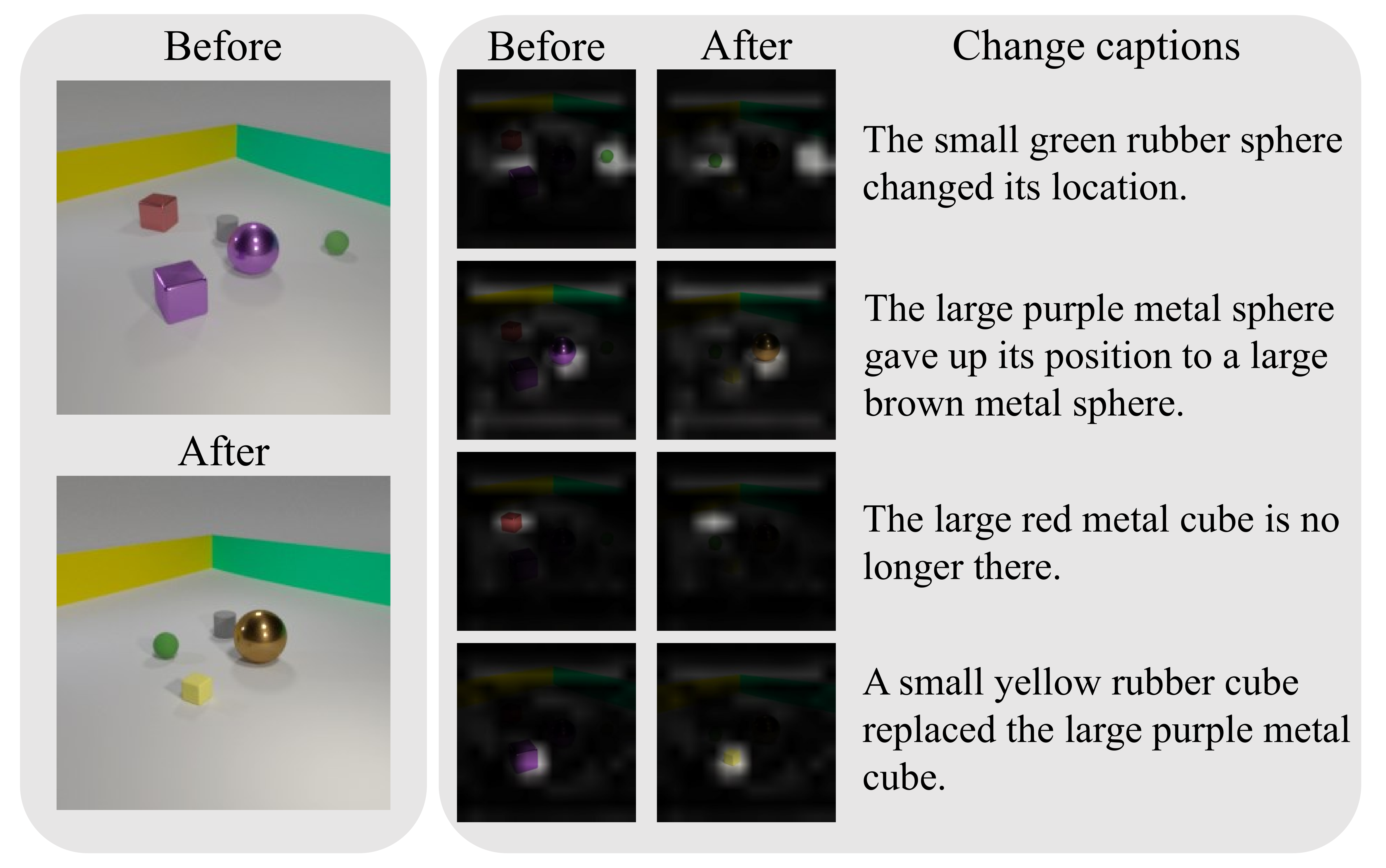}
\caption{Given two images of a scene observed before and after multiple changes \cG{(the first column)}, we generate change captioning for each scene change along with attention maps \cG{(the second and third columns)} indicating the region of changed objects.}
\label{fig:figure1}
\end{figure}

\cB{Several methods have been proposed for the change captioning~\cite{jhamtani2018learning,park2019robust,shi2020finding,oluwasanmi2019fully,oluwasanmi2019captionnet}.} \cB{Most} existing works focus on describing a single change. As a practical matter, multiple changes could be manifested within an image pair. Jhamtani and Berg-Kirkpatrick \cite{jhamtani2018learning} studied scene change captioning with multiple changes; however, they addressed the problem with a known number of changes which is not provided in real-world problems. Practically, detecting and describing scene changes without prior information regarding changes is more useful in terms of providing information to users. \cB{We address multi-change captioning, where changed regions are localized and language descriptions of scene changes are generated from pair of images with an unknown number of changes, as shown in Figure~\ref{fig:figure1}.}

Change captioning requires capturing relationships between \cB{image pairs}, localizing changed regions, and generating language descriptions. In this work, we introduce a simple but effective framework Multi-Change Captioning transformers (MCCFormers) based on encoder-decoder transformer \cite{vaswani2017attention} which performs well in natural language processing. The encoder transformer captures relationship between local regions in two images to detect scene changes. Then, the decoder transformer attends over changed regions and generates language descriptions of the changes. In contrast to existing methods that generate a static attention map \cite{park2019robust,shi2020finding}, the decoder transformer changes the spatial attention for each generated word. Consequently, the decoder transformer can distinguish between different changes and avoid confusing them for one another.

To evaluate multi-change captioning and localization ability, we build a novel CLEVR-Multi-Change dataset consisting of image pairs containing multiple changes, change captions, and bounding boxes of the changed region. We compare the proposed MCCFormers model with several state-of-the-art methods under the multi-change setup. The experimental results show that the proposed method performed well in both change captioning and localization.

The contributions of our work are three-fold: (i) We address a novel task of multi-change captioning and propose a dataset for this task where multiple changes exist in before- and after-change images and the number of changes is unknown; (ii) We propose MCCFormers, which consists of encoder-decoder transformers that capture relationships between images and densely correlates image regions with words; (iii) The proposed MCCFormers outperforms existing methods in terms of four conventional image captioning evaluation metrics and shows promising ability on localization for multi-sentence change captioning.

\section{Related Work}

\subsection{Change Detection}

Change detection from scenes captured from different moments has been studied in various research fields.  \cite{herbst2011toward,ambrucs2014meta,halber2019rescan} discussed change detection from indoor scenes. There are also existing studies which discuss change detection for disaster management \cite{sakurada2015change,fujita2017damage}, resource monitoring \cite{khan2017forest,saha2019unsupervised}, and vehicle navigation \cite{alcantarilla2018street}. \cB{Among existing studies}, \cite{herbst2011toward,ambrucs2014meta,halber2019rescan} proposed rule-based methods for detecting changed parts from a set of 3-D maps. \cite{khan2017forest,saha2019unsupervised,alcantarilla2018street,sakurada2015change,fujita2017damage} discuss generating pixel-level maps for indicating the changed region between image pairs. \cB{Instead of change detection,} we address localizing and describing changes.

\subsection{Image Captioning}
Image captioning is a \cB{well studied} topic at the intersection of computer vision and natural language processing. Vinyals \etal \cite{vinyals2015show} proposed encoder-decoder architecture where an encoder extracts image features and a decoder generates a description of an image. Xu \etal \cite{xu2015show} introduced the attention mechanism to align each word and relevant region in an image. Inspired by the human visual system, Anderson \etal proposed a combined bottom-up and top-down attention mechanism \cite{anderson2018bottom}. Following the success of transformers \cite{vaswani2017attention} in natural language processing, transformer-based approaches have been introduced for image captioning \cite{cornia2020meshed,li2019entangled,herdade2019image}. Different from image captioning \cB{for a single image}, we address change captioning which requires capturing the relationship between two images.

\subsection{Change Captioning}
Several studies have focused on change captioning which describes a change between two images from different moments. The Spot-the-Diff dataset which consists of 13,192 scene change image pairs was constructed by Jhamtani and Berg-Kirkpatrick \cite{jhamtani2018learning}. Each image pair has 1.86 change description sentences on average. However, they addressed the problem with a known number of changes. By contrast, we study the multi-change captioning task in which the number of scene changes is not given. CLEVR-Change dataset was introduced by Park \etal \cite{park2019robust} to overcome several limitations of the Spot-the-Diff dataset including lack of viewpoint change and localization ground truth. \cG{The authors of \cite{qiu20203d} and \cite{qiu2020indoor} discussed change captioning from image pairs observed from multiple viewpoints.} This work addresses multi-change captioning with an unknown number of changes and we develop CLEVR-Multi-Change dataset to evaluate localization ability as well as captioning.

Jhamtani and Berg-Kirkpatrick \cite{jhamtani2018learning} proposed DDLA which computes a pixel-level difference between image pairs, limiting the ability for situations with viewpoint changes. By contrast, DUDA \cite{park2019robust} utilizes feature-level differences to enhance the robustness to viewpoint change. Similarly, the Siamese difference captioning model was proposed by Oluwasanmi \etal \cite{oluwasanmi2019fully,oluwasanmi2019captionnet}. M-VAM \cite{shi2020finding} separates viewpoint changes from semantic changes by evaluating the similarity of different patches of image pairs. Spatial attentions used in DUDA and M-VAM are static, which limits their ability to distinguish different changes.

In this work, we build transformer-based encoder-decoder models for multi-change captioning. The encoder transformer computes patch-level similarity with multi-head attention which captures different types of changes between image pairs. The decoder transformer performs multi-head attention over image patches from the encoder, which captures the relation between generated words and image regions and thus can distinguish different changes.

\begin{figure}
\centering
\includegraphics[width=\linewidth]{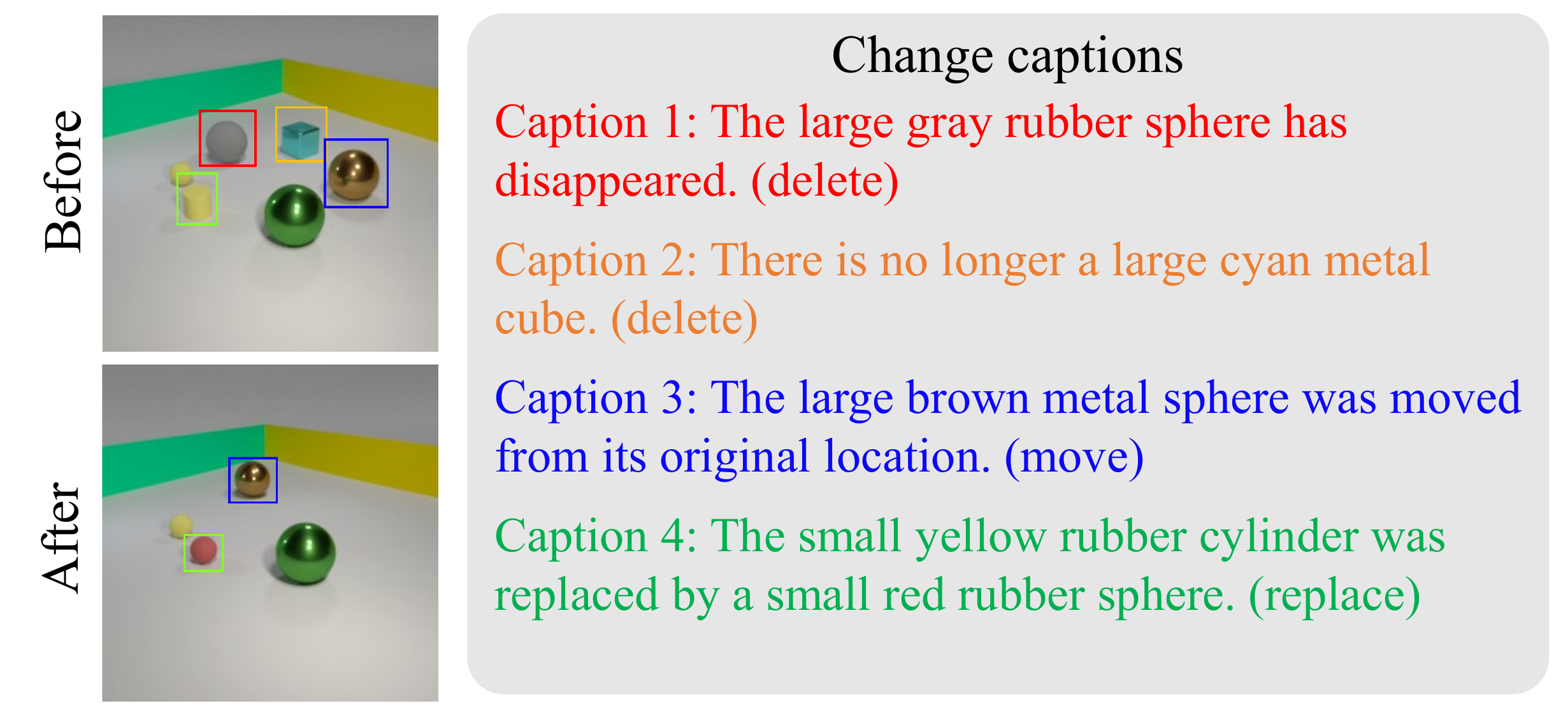}
\caption{Example from the CLEVR-Multi-Change dataset. The changed objects are highlighted by bounding boxes with the same color as the associated change captions. More dataset examples can be found in the supplementary material.}
\label{fig:figure2}
\end{figure}

\section{CLEVR-Multi-Change Dataset}

Existing change captioning studies mainly focus on single changes. 
However, identifying and distinguishing multiple change regions simultaneously manifested in image pairs is necessary due to frequent human activities. Moreover, change region localization is also critical in a variety of applications. For example, localizing target objects is essential for robot manipulation applications. To address these issues, we propose the CLEVR-Multi-Change dataset for diagnosing the ability of change localization and captioning in image pairs involving multiple changes based on the CLEVR engine \cite{johnson2017clevr} and CLEVR-Change dataset \cite{park2019robust}.

\textbf{Image Pairs Generation.} To generate a variety of scenes, we place objects with random shapes (cube, sphere, cylinder), colors (red, blue, yellow, green, brown, cyan, gray, purple), sizes (large, small), and materials (metal, rubber) into a simulated environment. We considered four atomic change types, namely ``add", ``delete", ``move", and ``replace" an object.
We set a virtual camera to create image pairs by observing a scene before and after scene change operations. We also add a random position change to cameras.
We generate each scene consisting of one to four changes within image pairs. We also record bounding boxes of changed objects for localization evaluation. 

\begin{table}
\begin{center}
\scalebox{0.74}[0.74]{%
\centering
\begin{tabular}{l|c c c c}
\toprule[1pt]
\multicolumn{1}{l|}{Dataset}&
\multicolumn{1}{c}{Multi-}&
\multicolumn{1}{c}{Viewpoint}&
\multicolumn{1}{c}{Localization}&
\multicolumn{1}{c}{Total}\\

& change & change & & image pairs \\ 

\midrule[0.5pt]

Spot-the-Diff \cite{jhamtani2018learning} & $\checkmark$ & & & 13,192\\
CLEVR-Change \cite{park2019robust} &   & $\checkmark$ & $\checkmark$ & 79,606\\
CLEVR-Multi-Change & $\checkmark$ & $\checkmark$ &  $\checkmark$ & 60,000\\

\bottomrule[1pt]

\end{tabular}
}
\end{center}
\caption{Change captioning dataset comparison.}\label{table1}
\end{table}

\begin{table}
\begin{center}
\scalebox{0.87}[0.87]{%
\begin{tabular}{l|c c c c}
\toprule[1pt]
\multicolumn{1}{l}{}&
\multicolumn{1}{|c}{1 Change}&
\multicolumn{1}{c}{2 Changes}&
\multicolumn{1}{c}{3 Changes}&
\multicolumn{1}{c}{4 Changes}\\
\midrule[0.5pt]
Image Pairs &  15,137  &  14,873 &  14,988 &  15,002  \\
Captions &  75,685 & 74,365  &  74,940 & 75,010 \\
Bboxes & 22,775 & 44,495 & 67,329 & 89,744 \\
\bottomrule[1pt]

\end{tabular}
}
\end{center}
\caption{CLEVR-Multi-Change dataset statistics.}\label{table2}
\end{table}

\begin{figure}
\centering
\includegraphics[width=\linewidth]{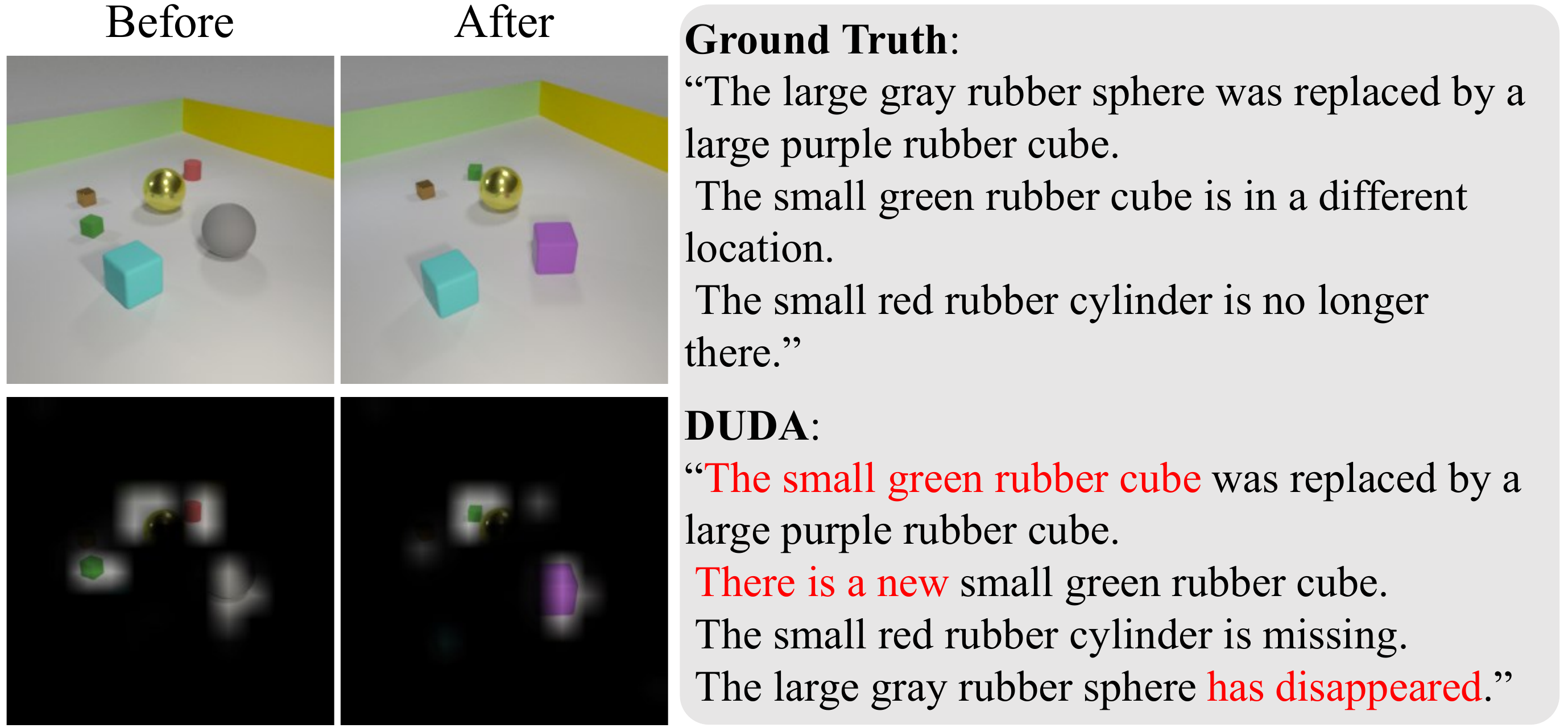}
\caption{Preliminary study on DUDA for describing multiple changes. Incorrect captions are in \cR{red} font.}
\label{fig:figure3}
\end{figure}

\begin{figure*}
\centering
\includegraphics[width=\linewidth]{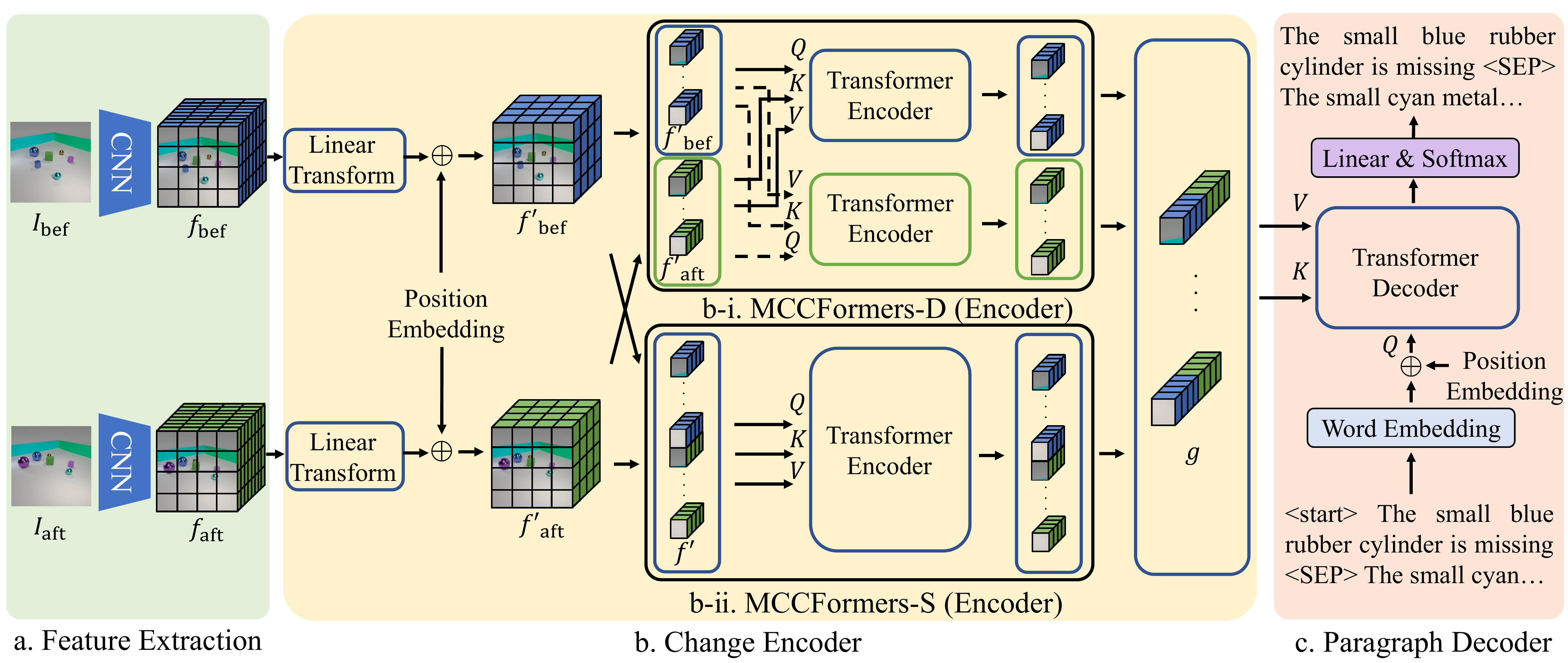}
\caption{Overall framework of MCCFormers: (a) Image features are extracted using CNN. We experimented with two encoders: (b-i) MCCFormers-D (Encoder) and (b-ii) MCCFormers-S (Encoder). We then feed representations from the encoder to the decoder (c) for caption generation. Best viewed in color.}
\label{fig:figure4}
\end{figure*}

\textbf{Eliminating Scene Change Ambiguity.} Unlike the CLEVR-Change dataset, we added two ``walls" with solid colors as background to reduce ambiguous correspondences between \cB{images due to the lack of camera information}. We deleted \cB{descriptions of} object relationships within an image (\eg to the left of) to prevent the ill-posed problem. To eliminate the ambiguity of change combinations (\eg ``replace a red cube with a blue cylinder" equals ``delete a red cube" and then ``add a blue cylinder"), we restrict the maximum change number to one for every object and region.

\textbf{Caption Generation.} The change captions are automatically generated based on recorded scene change information and pre-defined change sentence templates. We create five captions for each image pair with different sentence templates. The sentence order is randomly determined for two-, three-, and four-change image pairs.

We show the comparison with two extant datasets in Table~\ref{table1}. Statistics of our dataset and example are provided in Table~\ref{table2} and Figure~\ref{fig:figure2}, respectively. We split the dataset into 2/3, 1/6, and 1/6 for training, validation, and testing, respectively. We refer to the supplementary material for more details about the dataset generation process.

\textbf{Task Definition.} Given images of before and after changes $I_{\rm{bef}}$ and $I_{\rm{aft}}$, there are $N$ scene changes between them. Now, we define $S^i (i\in[1,\cdots,N])$ as a description of the $i^{th}$ change consisting of a sequence of words ($w_1^i,\cdots,w_M^i)$ with a maximum length of $M$. The multi-change captioning task aims to generate all $S^i$ from $I_{\rm{bef}}$ and $I_{\rm{aft}}$ with an unknown number of changes $N$. We consider predicting all sentences as a single sequence such as ($w_1^1,\cdots,w_M^1,<SEP>,w_1^2\cdots$). In our proposed framework, spatial attention is dynamically associated with each word and therefore localization of each change can be computed by averaging the attention maps of each word.

\textbf{Preliminary Study.} We evaluated a previous state-of-the-art method, DUDA, on this dataset and one example result is shown in Figure~\ref{fig:figure3}. Although DUDA can determine the changed region, \cB{it is confused} by multiple changes and generates change captions which are only partially correct. For example, DUDA generated ``The small green rubber cube was replaced by a large purple rubber cube", but the ground truth is ``The large gray rubber sphere was replaced by a large purple rubber cube", which suggests that DUDA attended wrong objects mentioned in different ground truth.

A dense correlation between different image regions of before- and after-change images is necessary for identifying changes. Furthermore, to distinguish and generate captions for each change, the correlation between change regions and sentences is critical.

\section{Approach}

Figure~\ref{fig:figure4} shows the proposed Multi-Change Captioning transformers (MCCFormers). Given two images ($I_{\rm{bef}}$ and $I_{\rm{aft}}$) of \cB{before and after multiple changes, MCCFormers generates a paragraph of descriptions of changes in image pairs}. Following existing methods like DUDA \cite{park2019robust} and M-VAM \cite{shi2020finding}, we first extract image features $f_{\rm{bef}}$ and $f_{\rm{aft}}$ using the CNN structure. We then feed the features to the transformer \cite{vaswani2017attention}-based encoder-decoder model. A transformer encoder densely correlates each image patch of before- and after-change image pairs and a decoder further correlates each word with image patches for generating \cB{descriptions} of multiple changes.

\subsection{Change Encoder}

It is necessary to distinguish and separate different change regions in scenes involving multiple changes, requiring a dense correlation of different regions between image pairs. To obtain the relationships of different image patches in image pairs, a mechanism to compare and correlate each image patch between image pairs is required. M-VAM correlates feature pairs by introducing an inner production operation of features. Compared with inner-production, the multi-head attention mechanism introduced in transformer-based encoders computes multiple types of attentions to correlate different patches. Thus, we consider adopting transformer-based encoders.

Different from recent transformer-based models for computer vision tasks such as DETR \cite{carion2020end} which takes a single image as its input, change captioning tasks feed two images.
Given image feature pairs $f_{\rm{bef}}$ and $f_{\rm{aft}}$ with dimension $\mathbb{R}^{W \times H \times D}$ (where W, H, and D are, respectively, the width, height, and channel of features), we consider two variants of encoder: Multi-Change Captioning transformers-Dual (MCCFormers-D) and Multi-Change Captioning transformers-Single (MCCFormers-S) (Figure~\ref{fig:figure4} (b-i) and (b-ii), respectively). For both variants, we first transform $f_{\rm{bef}}$ and $f_{\rm{aft}}$ to $f'_{\rm{bef}}$ and $f'_{\rm{aft}}$ with dimension $\mathbb{R}^{W \times H \times d_{encoder}}$. To accomplish this, we use linear transformation and add a position embedding as follows:

$$
f'(x,y) = W_{lt} f(x,y) + b_{lt} + pos(x,y) \eqno{(1)}
$$
\noindent
where $W_{lt}$ and $b_{lt}$ are learnable parameters of linear transformation and $pos(x,y)$ is learnable position embedding.

\textbf{MCCFormers-D.} In this variant, we use two transformer encoders with shared weights. To capture relevance between local regions of two images, we employ the co-attention mechanism \cite{lu2019vilbert}. In contrast to the original co-attention mechanism which takes linguistic tokens and object proposals from an image as input, we consider a set of patches of before and after images as input. Given two feature maps from before and after images $f'_{\rm{bef}}$ and $f'_{\rm{aft}}$, we consider that the query feature is either from before- or after-change images and the key and value feature is from the other one. After processing with the encoder with $N_e$ layers, we concatenate the output from features $g_{\rm{bef}}$ and $g_{\rm{aft}}$ over feature dimension as $g\in \mathbb{R}^{W \times H \times 2d_{encoder}}$.

\textbf{MCCFormers-S.} Different from
MCCFormers-D, we use MCCFormers-S to capture image patch relationships among both inter- and intra-image pairs. We first concatenate $f'_{\rm{bef}}$ and $f'_{\rm{aft}}$ to $f' \in \mathbb{R}^{2W\times H \times d_{encoder}}$. We then pass $f'$ to the standard transformer structure, which is similar to the BERT model \cite{devlin2019bert} that takes a sequence of two sentences as input. Following MCCFormers-D, the feature maps are converted to $g\in \mathbb{R}^{W \times H \times 2d_{encoder}}$. Compared with MCCFormers-D, which only considers relevance between image pairs, this structure also captures image patches' relevance within both before- and after-change images.

\subsection{Paragraph Decoder}

In the multi-change captioning task, due to the coexistence of multiple changes, it is critical to distinguish different change regions and dynamically attend to different regions during the generation of different sentences. Transformer decoders accomplish this by attending to information from different patches during the generation process.

Therefore, we adopt a standard transformer decoder for generating captions. We first use a word embedding layer to transfer input sentences and add a learnable position embedding. Next, the sentence features are processed through a masked self-attention and feed-forward network. The cross-attention between a sentence and the encoder's output features is then computed and further processed by a feed-forward layer. The decoder layer is iterated for $ N_d$ layers.

The transformer decoder computes attention over image features for each word during the sentence generation process. Thus, image attention can be computed for every sentence by averaging the attention maps of each word in a sentence. In contrast, DUDA and M-VAM compute a single spatial attention map for an entire paragraph.

\subsection{Learning Process}

From the input of images ($I_{\rm{bef}}$ and $I_{\rm{aft}}$) observed before and after scene changes, the decoder generates a word sequence with length $T$. We denote the target sequence as $(w_1^{*}, ..., w_T^{*})$. We adopt a cross-entropy loss for network training, where  $\theta$ indicates the learnable parameter:

$$
L_{XE} = \sum_{t=1}^{T} -{\rm{log}}(p_{\theta}(w_t^{*} \vert (w_1^{*},  ..., w_{t-1}^{*}),I_{\rm{bef}}, I_{\rm{aft}})) \eqno{(2)}
$$

\begin{table}
\begin{center}
\scalebox{0.9}[0.9]{%
\begin{tabular}{l | c }
\toprule
\multicolumn{1}{l |}{Methods}&
\multicolumn{1}{c}{BLEU-4}\\

\midrule

DUDA \cite{park2019robust} & 76.1  \\
DUDA Encoder + Transformer Decoder & 79.1 \\
M-VAM \cite{shi2020finding}  & 62.9 \\

M-VAM Encoder + Transformer Decoder & 65.8 \\

\midrule
MCCFormers-D (concat. over patches) & 80.1 \\
MCCFormers-D (concat. over feature dimension) & 82.3 \\
MCCFormers-S (concat. over patches) & 80.6 \\
MCCFormers-S (concat. over feature dimension) & \textbf{83.3} \\

\bottomrule

\end{tabular}

}
\end{center}
\caption{BLEU-4 evaluation of different methods applied to the CLEVR-Multi-Change dataset. (concat. : concatenation)} \label{table3}

\end{table}

\begin{table*}
\begin{center}
\scalebox{0.9}[0.9]{%
\begin{tabular}{l | c c c c c c c c}
\toprule
\multicolumn{1}{l |}{Methods}&
\multicolumn{5}{c}{BLEU-4 \cite{papineni2002bleu}}&
\multicolumn{1}{c}{CIDEr}&
\multicolumn{1}{c}{METEOR}&
\multicolumn{1}{c}{SPICE} \\

& Overall & 1 Change & 2 Changes & 3 Changes & 4 Changes & \cite{vedantam2015cider} & \cite{banerjee2005meteor} & \cite{anderson2016spice} \\

\midrule

DUDA \cite{park2019robust} & 76.1 & 94.7 & 76.3 & 73.1 & 70.9 & 480.1 & 47.4  &  66.6 \\
M-VAM \cite{shi2020finding}  & 62.9  & 79.0 & 61.7 & 59.9 & 57.9 & 338.1 & 41.3 & 55.9 \\
\midrule
MCCFormers-D & 82.3 & \textbf{98.4}  & \textbf{82.9} &  79.2 & \textbf{80.6} & \textbf{539.3} & \textbf{52.1} & \textbf{71.7} \\
MCCFormers-S & \textbf{83.3}  & 96.5  &  82.6 &  \textbf{81.8} & 80.0 & 523.3 & 51.5 & 70.0 \\

\bottomrule

\end{tabular}
}
\end{center}
\caption{\cB{Results} on the CLEVR-Multi-Change dataset.} \label{table4}

\end{table*}

\begin{table}
\begin{center}
\scalebox{0.667}[0.667]{%
\begin{tabular}{l | c c c c c c}
\toprule
\multirow{2}{*}{Methods}&
\multicolumn{5}{c}{Accuracy}&
\multirow{2}{*}{MAE} \\
&All&1 change & 2 change & 3 change & 4 change & \\

\midrule

DUDA \cite{park2019robust} & 83.3  & 96.8 & 84.8 & 74.4 & 76.0 & 0.169 \\
M-VAM \cite{shi2020finding}  & 69.9 & 91.2 & 68.4 & 56.5 & 62.0 & 0.317 \\
\midrule
MCCFormers-D & \textbf{92.5}  & \textbf{99.4} & 94.8 & 81.9 & \textbf{93.9} & \textbf{0.075} \\
MCCFormers-S & 92.4 & 97.9 & \textbf{95.9} & \textbf{91.9} & 83.7 & \textbf{0.075} \\

\bottomrule

\end{tabular}

}
\end{center}
\caption{Sentence number accuracy (\%) and mean absolute error (MAE) on the CLEVR-Multi-Change dataset.} \label{table5}

\end{table}

\section{Experiments}
\subsection{Experimental Setup}

\textbf{Experiments and Datasets.} We conducted experiments on both multi-change and single change setups. \cG{We also implemented DUDA and M-VAM without models modification compared to the models proposed in original papers and evaluated the models performance on paragraph generation.} We also report the performance of the methods on two previous datasets i.e., Spot-the-Diff (containing multi- and single change setups, where we sampled instances within four changes for the multi-change setup) and the CLEVR-Change dataset (single change).

\textbf{Evaluation Metrics.} We adopt conventional image captioning and change captioning evaluation metrics for performance comparison: BLEU-4 \cite{papineni2002bleu}, CIDEr \cite{vedantam2015cider}, METEOR \cite{banerjee2005meteor}, and SPICE \cite{anderson2016spice}. These metrics evaluate the similarity of generated sentences with ground truth from different aspects. We also evaluated the accuracy in terms of the number of sentences in multi-sentence generation. We compute accuracy which measures whether the number of sentences of the generated sequence is correct and mean absolute error (MAE) to evaluate the difference in the number of sentences between ground truth and generated sequence. \cG{We prepared five ground truth paragraphs with different sentence orders for each image pair. Therefore, the sentence order has less influence on evaluation results.} To assess the localization ability in multi-change captioning, we introduce an evaluation metric based on the Pointing Game \cite{zhang2018top}. We record the bounding boxes for changed objects and transfer the obtained attention map to the original image size with bilinear interpolation. We then select the top-$K$ pixels with the largest values in attention maps and compute the detected change region (where the top-$K$ pixels are inside the bounding box of a changed region) over all of the changed regions in the ground truth. The overall accuracy is averaged over all changes contained in the test data. \cG{We set $K$ to 1 for add and delete and 2 for move and replace where the bounding boxes of objects might be different due to the scene change.}

\textbf{Implementation Details.} Similar to \cite{park2019robust,shi2020finding}, we use ResNet-101 \cite{he2016deep} pretrained on the ImageNet dataset \cite{deng2009imagenet} to extract image features from images with a $224 \times 224$ resolution. The obtained feature maps dimension is $14 \times 14 \times 1024$. 
We implemented transformers with two layers and four heads for both encoders and decoders.
The dimensions of input features to encoder $d_{encoder}$ and decoder $d_{decoder}$ are 512 and 1024, respectively. For the feedforward network, the dimensions are $4d_{encoder}$ and $4d_{decoder}$ for encoders and decoders, respectively.
We set the learning rate to 0.0001 and trained models for 40 epochs with the Adam optimizer \cite{kingma2014adam} during all implementations. 

\textbf{Baselines.} We use DUDA and M-VAM for comparison during experiments and two variants of the transformer network. We set the hidden state dimension for all LSTM structures to 512 in both DUDA and M-VAM (see the supplementary material for details of the implementation of DUDA and M-VAM).

\subsection{CLEVR-Multi-Change dataset}

\begin{figure*}
\centering
\includegraphics[width=\linewidth]{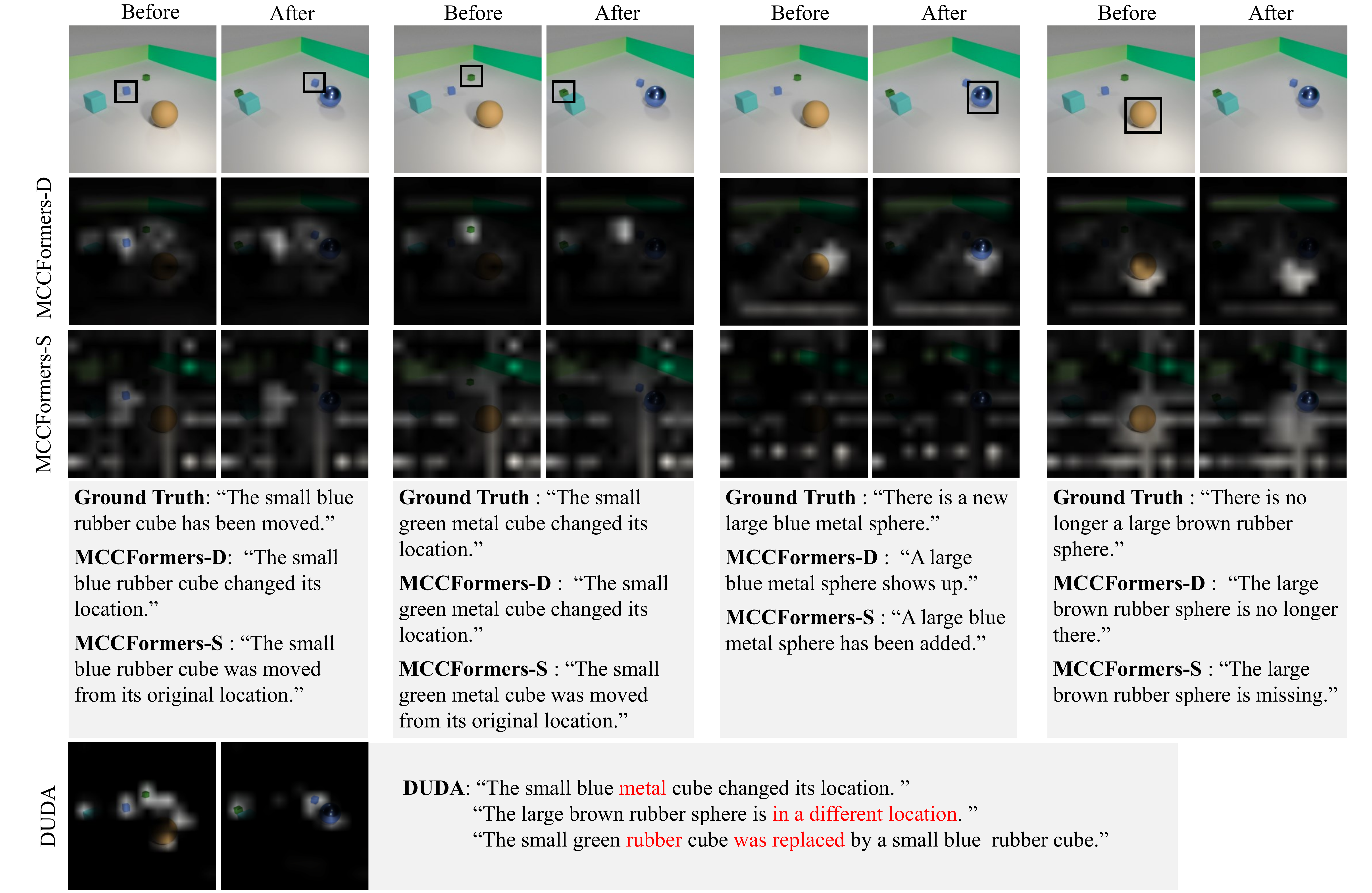}
\caption{Visualization of an example from the CLEVR-Multi-Change dataset. We show the attention maps for the proposed methods and DUDA and generated sentences. Incorrect captions are in \cR{red} font. We highlighted changed regions in black bounding boxes.}
\label{fig:figure5}
\end{figure*}

\textbf{Ablation Study.} We evaluated different structure choices of change encoders and decoders (Table~\ref{table3}). DUDA and M-VAM scored 76.1 and 62.9 for BLEU-4, respectively. We removed the sum operation over the spatial region of features (DUDA) and average pooling operations (M-VAM) which are applied before feeding features to decoders. We then replaced the decoders with transformer decoders. The use of transformer decoders improved the performance of both two methods. 

Next, we conducted experiments regarding MCCFormers. Before feeding features to the decoder, we concatenate features of before- and after-change images in two ways: concatenation operation over patches (input to decoder: $g \in \mathbb{R}^{2W\times H \times d_{encoder}}$) and over feature dimension (input to decoder: $g \in \mathbb{R}^{W \times H \times 2d_{encoder}}$). All methods outperformed previous methods in terms of BLEU-4. MCCFormers-D and MCCFormers-S with concatenation over feature dimension performed the best.
\cB{In the case of relatively small viewpoint change, patches of the same region in two images are concatenated, which could improve the effectiveness of the concatenation over feature dimension. We will investigate the robustness to viewpoint change (especially larger change) for future work.}
In the remaining experiments, we use MCCFormers-D and MCCFormers-S with concatenation over feature dimension.

\textbf{Sentence Generation.} Experimental results of the different evaluation metrics are shown in Table~\ref{table4}. Both MCCFormers-D and MCCFormers-S outperformed previous methods with respect to all metrics. MCCFormers-S obtained the highest BLEU-4 and outperformed previous methods by +7.2. 

For one-change instances, the differences between the proposed methods and previous methods are relatively small. For instances with multiple changes, the two proposed methods exhibited better robustness. The transformer encoder learns a dense correlation among all local regions of change image pairs, and the decoder model further correlates each word with image regions, making the models better at distinguishing different changes.

\textbf{Evaluation of Sentence Number Accuracy.} Results concerning sentence number accuracy and MAE are shown in Table~\ref{table5}. Similar to the results in Table~\ref{table4}, the proposed methods obtained higher accuracy for sentence numbers compared to previous methods and achieved promising results for distinguishing changes, with an accuracy of 92.5\% for the MCCFormers-D method. The MAE results show that the average errors of the number of sentences generated by all methods are less than 1.

All methods achieved the highest scores for one-change sentence and the MCCFormers-D model obtained 99.4\% accuracy. For two-, three- and four-change instances, the accuracy of previous methods was degraded while the two proposed methods show promising stability for scenes with multiple changes.

\begin{figure}
\centering
\includegraphics[width=\linewidth]{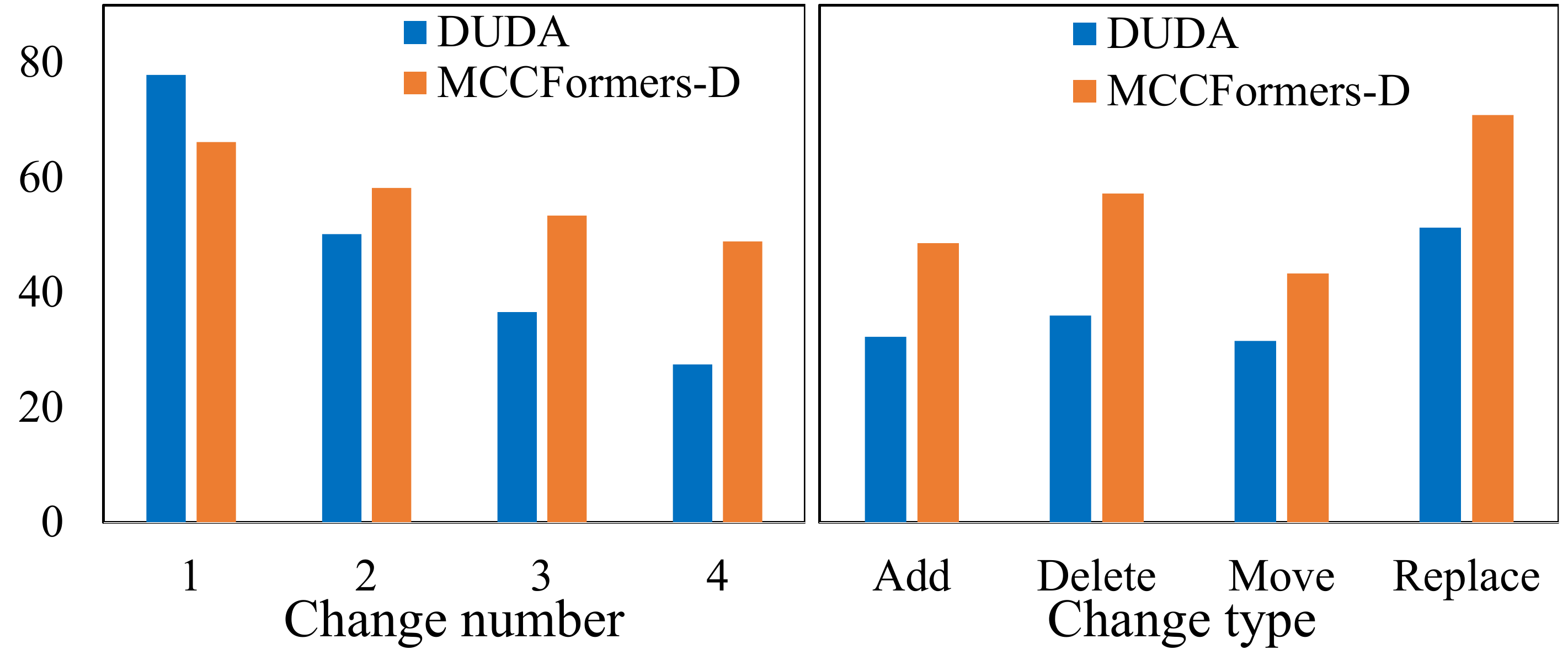}
\caption{Pointing Game accuracy (\%) over different change numbers (left) and types (right) on the CLEVR-Multi-Change dataset.}
\label{fig:figure6}
\end{figure}

\textbf{Qualitative Results.} We show one example result in Figure~\ref{fig:figure5}. This example contains four changes. DUDA predicted three changes and both variants of MCCFormers generated correct sentences in terms of both change number and change contents. In addition, two captions generated by DUDA contain incorrect change types whilst two proposed methods generated correct captions for each change.

DUDA generates a single attention map for each given image pair. Therefore, the network could be struggling to distinguish each changed part from the attention map, limiting its ability in multi-change understanding. MCCFormers-S generated attention maps that tend to focus on unrelated regions as well as changed object regions. This structure refers to the patches of inter- and intra-image pairs, which might weaken the interpretability of attention maps. MCCFormers-D obtained separate attention maps for each sentence and attended to change regions. We refer to the supplementary material for more example results.

\textbf{Pointing Game Evaluation for Attention Maps.} We evaluated the localization accuracy of MCCFormers-D and DUDA (Figure~\ref{fig:figure6}). Since DUDA generates a single pair of attention maps for each image pair, we use the same attention map to evaluate each change. MCCFormers-D obtained an overall accuracy of 53.9\% for change localization and 40.0\% for DUDA. In both methods, localization performance degrades with the increase in change number. DUDA obtained higher localization accuracy for one change and our methods outperformed DUDA for two, three, and four changes, indicating the effectiveness of MCCFormers-D for detecting multiple changes.

Among different change types, both methods obtained the highest accuracy for replace change and the lowest accuracy for move change. The move change relates to two image positions, making it challenging for localization.

\begin{table}
\begin{center}
\scalebox{0.85}[0.85]{%
\begin{tabular}{l | c c c c}
\toprule
\multicolumn{1}{l |}{Methods}&
\multicolumn{1}{c}{BLEU-4}&
\multicolumn{1}{c}{CIDEr}&
\multicolumn{1}{c}{METEOR}&
\multicolumn{1}{c}{SPICE}\\

\midrule

\multicolumn{5}{@{}c}{\makecell{ \hspace{0.8cm}   Multiple change (one to four changes)}}\\

\midrule

DUDA \cite{park2019robust} & 5.4 & 24.8 & \textbf{10.6} & 12.9 \\

\midrule
MCCFormers-D & \textbf{6.2} & \textbf{28.8} & 10.2 & \textbf{17.8} \\
MCCFormers-S & 5.8 & 18.2 & 10.5 & 10.1 \\

\midrule

\multicolumn{5}{@{}c}{\makecell{ \hspace{0.8cm}   Single change}}\\

\midrule

DUDA \cite{park2019robust} & 8.1 & 34.0 & 11.5 & -  \\
FCC \cite{oluwasanmi2019fully} & 9.9 & 36.8 & \textbf{12.9} & -  \\
SDCM \cite{oluwasanmi2019captionnet} & 9.8 & 36.3 & 12.7 & -  \\
DDLA \cite{jhamtani2018learning} & 8.5 & 32.8 & 12.0 & -  \\
M-VAM \cite{shi2020finding} & 10.1 & 38.1 & 12.4 & 14.0  \\
M-VAM + RAF \cite{shi2020finding} & \textbf{11.1} & 42.5 & \textbf{12.9} & 17.1  \\

\midrule
MCCFormers-D & 10.0 & \textbf{43.1} & 12.4 & \textbf{18.3} \\
MCCFormers-S & 9.8 & 41.6 & 12.3  & 16.3 \\

\bottomrule

\end{tabular}
}
\end{center}

\caption{\cB{Results} on the Spot-the-Diff dataset.}\label{table6}
\end{table}

\begin{figure}
\centering
\includegraphics[width=\linewidth]{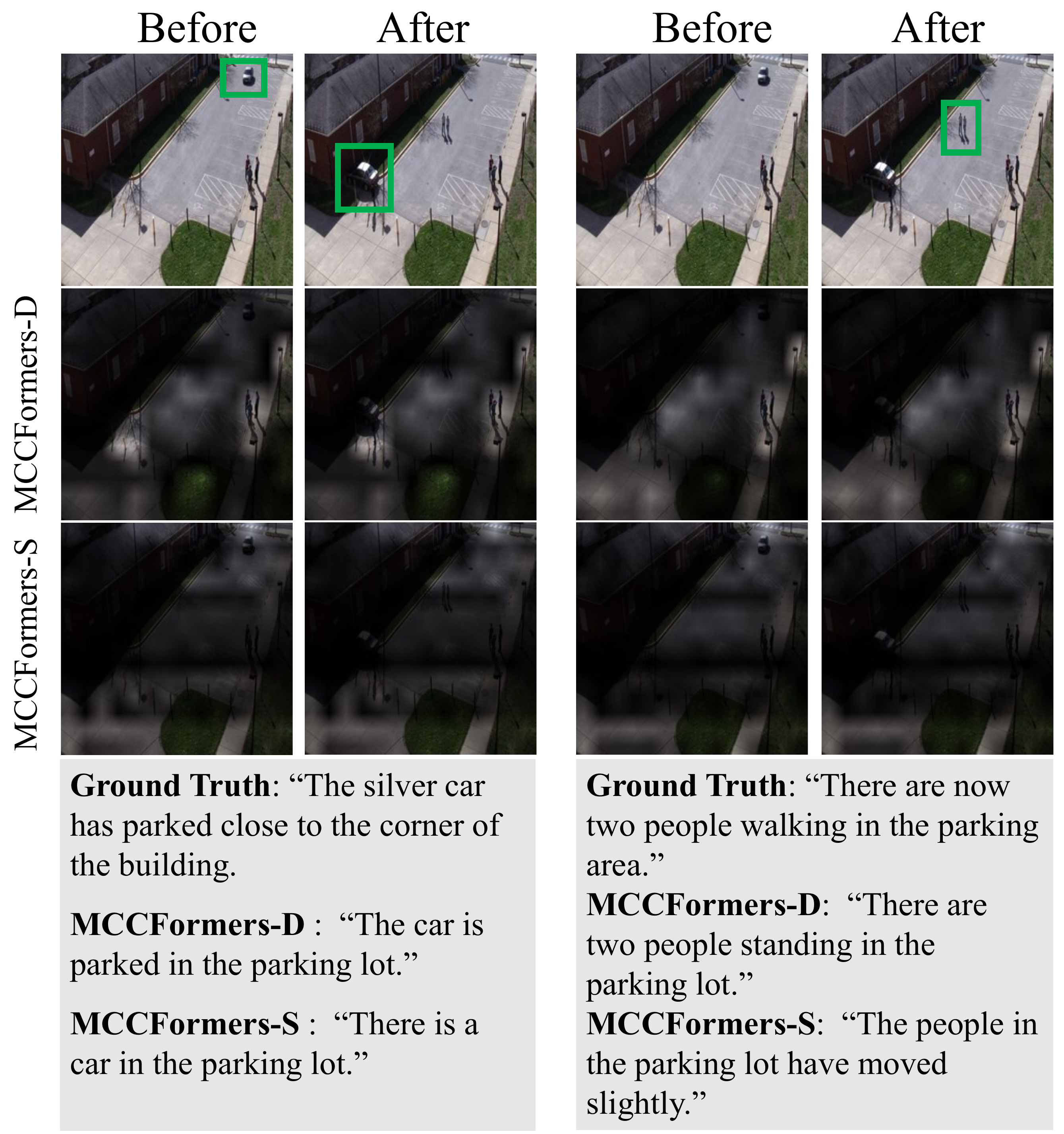}
\caption{Visualization of an example of proposed methods on the Spot-the-Diff dataset (multiple change). We highlighted changed regions in green bounding boxes.}
\label{fig:figure7}
\end{figure}

\subsection{Spot-the-Diff Dataset}

Spot-the-Diff contains multiple changes within image pairs. We first extract all instances containing one to four changes and report the results of DUDA and MCCFormers in Table~\ref{table6} (top three rows). For this dataset, MCCFormers obtained comparable results with DUDA. We further show one example in Figure~\ref{fig:figure7}. For the example containing two changes, the two methods correctly generated two related sentences. However, compared to ground truth sentences, the generated captions lack detailed attribute information, such as the color of the car is ``silver" and the detailed location is the ``corner of the building". We also found that for MCCFormers-D, the generated attention maps plausibly attended related regions but highlighted many unrelated image regions. MCCFormers-S failed to point out the related regions and struggled at recognizing detailed changes.

We show the results of the single change setup in Table~\ref{table6} (bottom eight rows). The proposed methods obtained scores comparable to the state-of-the-art method M-VAM.

Compared to the proposed dataset, the Spot-the-Diff dataset contains fewer images, which might limit the performance of the transformer-based methods because they tend to require a large amount of training data. Further exploration of change captioning on large-scale real-world imagery is warranted in future research.

\begin{table}
\begin{center}
\scalebox{0.86}[0.86]{%
\begin{tabular}{l | c c c c}
\toprule
\multicolumn{1}{l |}{Methods}&
\multicolumn{1}{c}{BLEU-4}&
\multicolumn{1}{c}{CIDEr}&
\multicolumn{1}{c}{METEOR}&
\multicolumn{1}{c}{SPICE}\\

\midrule

DUDA \cite{park2019robust} & 47.3 & 112.3 & 33.9 & 24.5  \\
M-VAM \cite{shi2020finding} & 50.3 & 114.9 & 37.0  & 30.5  \\
M-VAM + RAF \cite{shi2020finding} & 51.3 & 115.8 & 37.8 & 30.7  \\
\midrule
MCCFormers-D & 52.4 & 121.6 & 38.3 & 26.8 \\
MCCFormers-S & \textbf{57.4} & \textbf{125.5} & \textbf{41.2}  & \textbf{32.4} \\

\bottomrule

\end{tabular}
}
\end{center}

\caption{\cB{Results} on the CLEVR-Change dataset.}\label{table7}
\end{table}

\subsection{CLEVR-Change Dataset}

We compare the different methods on the previous single change dataset CLEVR-Change in Table~\ref{table7}. The CLEVR-Change dataset requires understanding object relationships inside each image \cG{(\eg in front of) which are not included in the proposed dataset.} Therefore, compared with MCCFormers-D, MCCFormers-S obtained the highest scores as MCCFormers-S can capture relationships between image patches within the same image. MCCFormers outperformed the previous methods on this dataset in terms of most evaluation metrics, indicating the ability of proposed structures to correlate different regions in change image pairs and further connect the change region information with words in sentences. 

\section{Conclusion}
\cB{In this paper, we propose a novel multi-change captioning task and CLEVR-Multi-Change dataset for this task. To address the novel task, we proposed a transformer-based framework MCCFormers that densely correlates different image regions in image pairs and words. MCCFormers achieved state-of-the-art performance on both multi- and single-change captioning datasets, indicating the effectiveness of MCCFormers for change captioning tasks.
}

\section*{Acknowledgements}
We want to thank Yoshitaka Ushiku, Seito Kasai, Hikaru Ishitsuka, and Tomomi Satoh for their helpful comments during research discussions. This paper is based on results obtained from a project, JPNP20006, commissioned by the New Energy and Industrial Technology Development Organization (NEDO). Computational resource of AI Bridging Cloud Infrastructure (ABCI) provided by National Institute of Advanced Industrial Science and Technology (AIST) was used.

\section*{Supplementary Material}
This supplementary material provides additional implementation details of methods used in this article, including the proposed methods MCCFormers and previous methods, DUDA, and M-VAM. We also give a more detailed introduction of the CLEVR-Multi-Change dataset, including the details of the caption generation process and additional dataset examples. Additional experimental results on the CLEVR-Multi-Change dataset can be found in the last section of this material.

\setcounter{section}{0}
\renewcommand\thesection{\Alph{section}}

\section{Additional Implementation Details}

\begin{figure}
\centering
\includegraphics[height=14cm]{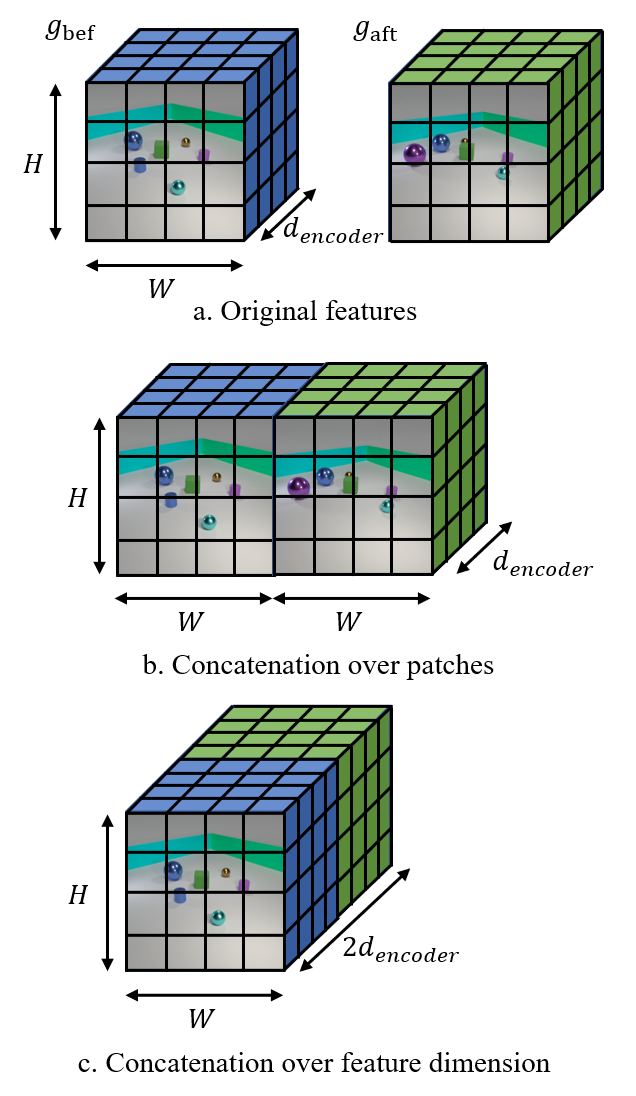}

\caption{Visualization of feature concatenation of encoder outputs.}
\label{fig:concat}
\end{figure}

\textbf{Feature Concatenation of MCCFormers.} Here, we provide more details of the feature concatenation operation used in the Ablation Study of subsection 5.2 and Table 3 in the main paper. The MCCFormers-D (encoder) outputs $g_{\rm{bef}}$ and $g_{\rm{aft}}$ with dimension of $\mathbb{R}^{W \times H \times d_{encoder}}$, respectively. The MCCFormers-S (encoder) outputs a feature map with dimension of $\mathbb{R}^{2W \times H \times d_{encoder}}$. We then separate the output to $g_{\rm{bef}}$ and $g_{\rm{aft}}$ with dimension of $\mathbb{R}^{W \times H \times d_{encoder}}$. For both two MCCFormers, we consider two ways to concatenate $g_{\rm{bef}}$ and $g_{\rm{aft}}$ (Figure~\ref{fig:concat} (a)): concatenation over patches (Figure~\ref{fig:concat} (b)) and concatenation over feature dimension (Figure~\ref{fig:concat} (c)), before feeding features to decoders. The experimental results are given in Table 3 of the main paper.

\textbf{DUDA.} We implemented DUDA based on the code \footnote{The implementation code of DUDA: \url{https://github.com/Seth-Park/RobustChangeCaptioning}} provided by the authors of DUDA. We set the dimension of the encoder and LSTM hidden layer of DUDA to 512.

\textbf{M-VAM.} We implemented M-VAM following the approach introduced in the original paper of M-VAM \cite{shi2020finding}. For encoder of M-VAM, two scalars in Equation (3) in \cite{shi2020finding} are learned during training. Regarding the sentence decoder, two LSTM with hidden state dimensions of 512 are trained. The network is trained with cross-entropy loss in an end-to-end manner.

For the implementation of DUDA and M-VAM, we used the same input image features, learning rate, optimizer, learning iteration as the proposed methods introduced in Implementation Details of subsection 5.1 of the main paper.

\begin{table*}
\begin{center}
\scalebox{0.99}[0.99]{%
\begin{tabular}{l | l }
\toprule
\multicolumn{1}{l |}{Change type}&
\multicolumn{1}{l}{Caption templates}\\

\midrule

Add  & ``A $<$s$>$ $<$c$>$ $<$t$>$ $<$z$>$ has been added.'' \\   
& ``A $<$s$>$ $<$c$>$ $<$t$>$ $<$z$>$ shows up.''  \\
& ``There is a new $<$s$>$ $<$c$>$ $<$t$>$ $<$z$>$.''  \\
& ``A new $<$s$>$ $<$c$>$ $<$t$>$ $<$z$>$ is visible.''  \\
& ``Someone added a $<$s$>$ $<$c$>$ $<$t$>$ $<$z$>$.''  \\

\midrule
Delete  & ``The $<$s$>$ $<$c$>$ $<$t$>$ $<$z$>$ has disappeared.''  \\
& ``The $<$s$>$ $<$c$>$ $<$t$>$ $<$z$>$ is no longer there.''  \\
& ``The $<$s$>$ $<$c$>$ $<$t$>$ $<$z$>$ is missing.''  \\
& ``There is no longer a $<$s$>$ $<$c$>$ $<$t$>$ $<$z$>$.''  \\
& ``Someone removed the $<$s$>$ $<$c$>$ $<$t$>$ $<$z$>$.''  \\

\midrule
Move  & ``The $<$s$>$ $<$c$>$ $<$t$>$ $<$z$>$ changed its location.''  \\
& ``The $<$s$>$ $<$c$>$ $<$t$>$ $<$z$>$ is in a different location.''  \\
& ``The $<$s$>$ $<$c$>$ $<$t$>$ $<$z$>$ was moved from its original location.''  \\
& ``The $<$s$>$ $<$c$>$ $<$t$>$ $<$z$>$ has been moved.''  \\
& ``Someone changed location of the $<$s$>$ $<$c$>$ $<$t$>$ $<$z$>$.''  \\

\midrule
Replace  & ``The $<$s$>$ $<$c$>$ $<$t$>$ $<$z$>$ was replaced by a $<$s1$>$ $<$c1$>$ $<$t1$>$ $<$z1$>$.''  \\
& ``A $<$s1$>$ $<$c1$>$ $<$t1$>$ $<$z1$>$ replaced the $<$s$>$ $<$c$>$ $<$t$>$ $<$z$>$.''  \\
& ``A $<$s1$>$ $<$c1$>$ $<$t1$>$ $<$z1$>$ is in the original position of $<$s$>$ $<$c$>$ $<$t$>$ $<$z$>$.''  \\
& ``The $<$s$>$ $<$c$>$ $<$t$>$ $<$z$>$ gave up its position to a $<$s1$>$ $<$c1$>$ $<$t1$>$ $<$z1$>$.''  \\
& ``Someone replaced the $<$s$>$ $<$c$>$ $<$t$>$ $<$z$>$ with a $<$s1$>$ $<$c1$>$ $<$t1$>$ $<$z1$>$.''  \\

\bottomrule

\end{tabular}

}
\end{center}
\caption{Caption templates used in CLEVR-Multi-Change dataset. $<$s$>$, $<$s1$>$: size; $<$c$>$, $<$c1$>$: color; $<$t$>$, $<$t1$>$: material; $<$z$>$, $<$z1$>$: shape.} \label{table8}

\end{table*}

\begin{table}
\begin{center}
\scalebox{0.95}[0.95]{%
\begin{tabular}{l | l l | c }
\toprule[1pt]
\multicolumn{1}{l}{Models}&
\multicolumn{1}{|l}{Layers}&
\multicolumn{1}{l}{Heads}&
\multicolumn{1}{|c}{BLEU-4 (Overall)}\\
\midrule[0.5pt]
\multirow{12}{*}{MCCFormers-D} &  \multirow{4}{*}{1}  &  1  & 59.0 \\
 &    &  2  & 65.8 \\
 &    &  4  & 71.0 \\
 &    &  8  & 76.8 \\ \cline{2-4}
 &  \multirow{4}{*}{2}  &  1  & 81.4 \\
 &    &  2  &  81.2 \\
 &    &  4  & 82.3 \\
 &    &  8  &  \textbf{82.5} \\\cline{2-4}
 &  \multirow{4}{*}{4}  &  1  & 60.3 \\
 &    &  2  & 59.8 \\
 &    &  4  & 64.8 \\
 &    &  8  & 77.2 \\
\midrule[0.5pt]
\multirow{12}{*}{MCCFormers-S} &  \multirow{4}{*}{1}  &  1  & 58.1 \\
 &    &  2  & 64.0 \\
 &    &  4  & 75.8 \\
 &    &  8  & 79.9 \\\cline{2-4}
 &  \multirow{4}{*}{2}  &  1  & 80.0 \\
 &    &  2  & 82.2 \\
 &    &  4  & \textbf{83.3} \\
 &    &  8  & 83.0 \\
\bottomrule[1pt]

\end{tabular}
}
\end{center}
\caption{BLEU-4 evaluation of different network designs (Layers and Heads) of MCCFormers on CLEVR-Multi-Change dataset.}\label{table9}
\end{table}

\section{Additional Details on CLEVR-Multi-Change Dataset}

\textbf{Caption Generation.} As introduced in section 3 of the main paper, the CLEVR-Multi-Change dataset consists of before- and after-change image pairs and captions that describe changes through language text. We record the change information during the generation of image pairs, including change type and attributes of related objects. The change captions are generated based on recorded change information and pre-defined sentence templates. 

All templates used in the CLEVR-Multi-Change dataset are shown in Table~\ref{table8}. The tags ``$<$s$>$ $<$c$>$ $<$t$>$ $<$z$>$'' and ``$<$s1$>$ $<$c1$>$ $<$t1$>$ $<$z1$>$'' in each template are instantiated during caption generation. For example, with the template ``A $<$s$>$ $<$c$>$ $<$t$>$ $<$z$>$ has been added.'' and an added object with attributes: small, red, metal, cube, the generated caption would be ``A small red metal cube has been added.''

\begin{figure*}
\centering
\includegraphics[width=\linewidth]{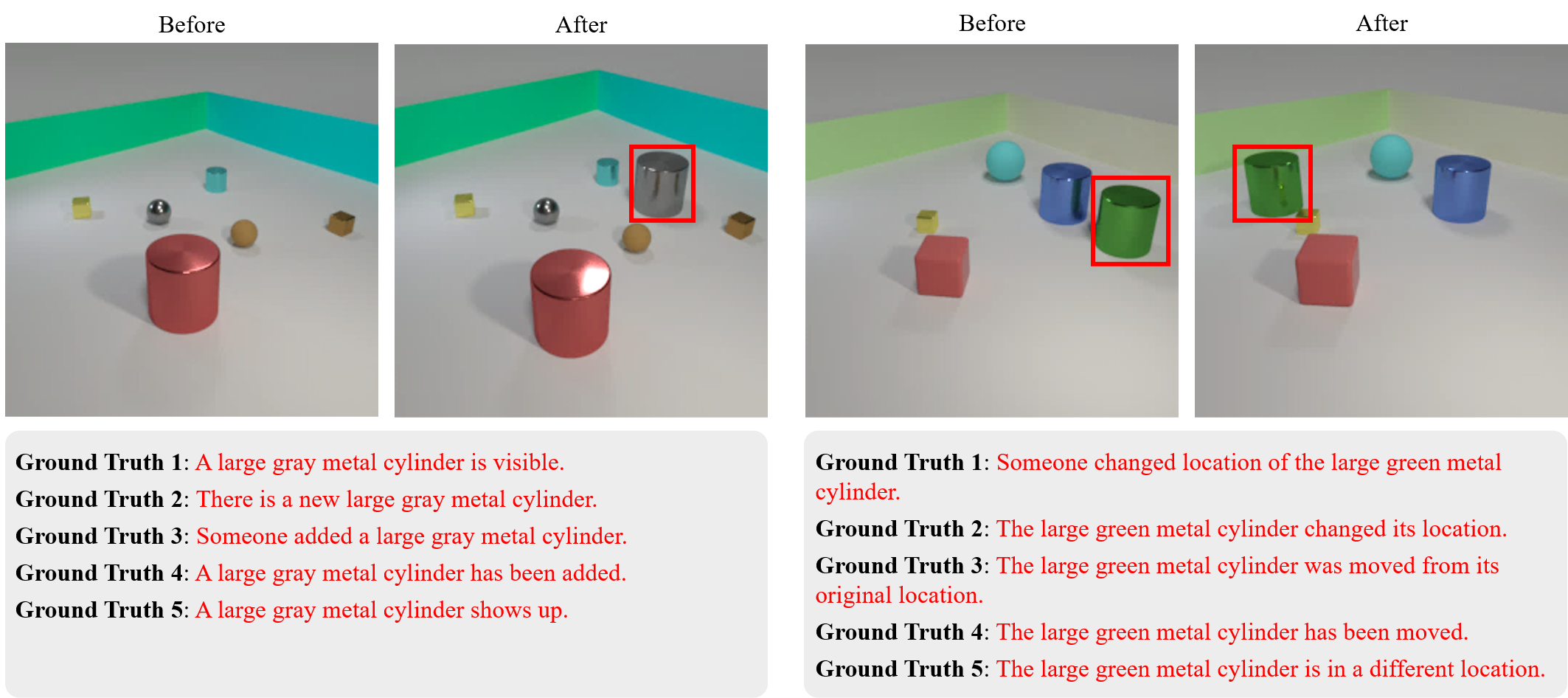}
\caption{One-change examples from the CLEVR-Multi-Change dataset. The changed objects are highlighted by rectangles with the same color as the associated change captions.}\label{fig:1change}
\end{figure*}

\begin{figure*}
\centering
\includegraphics[width=\linewidth]{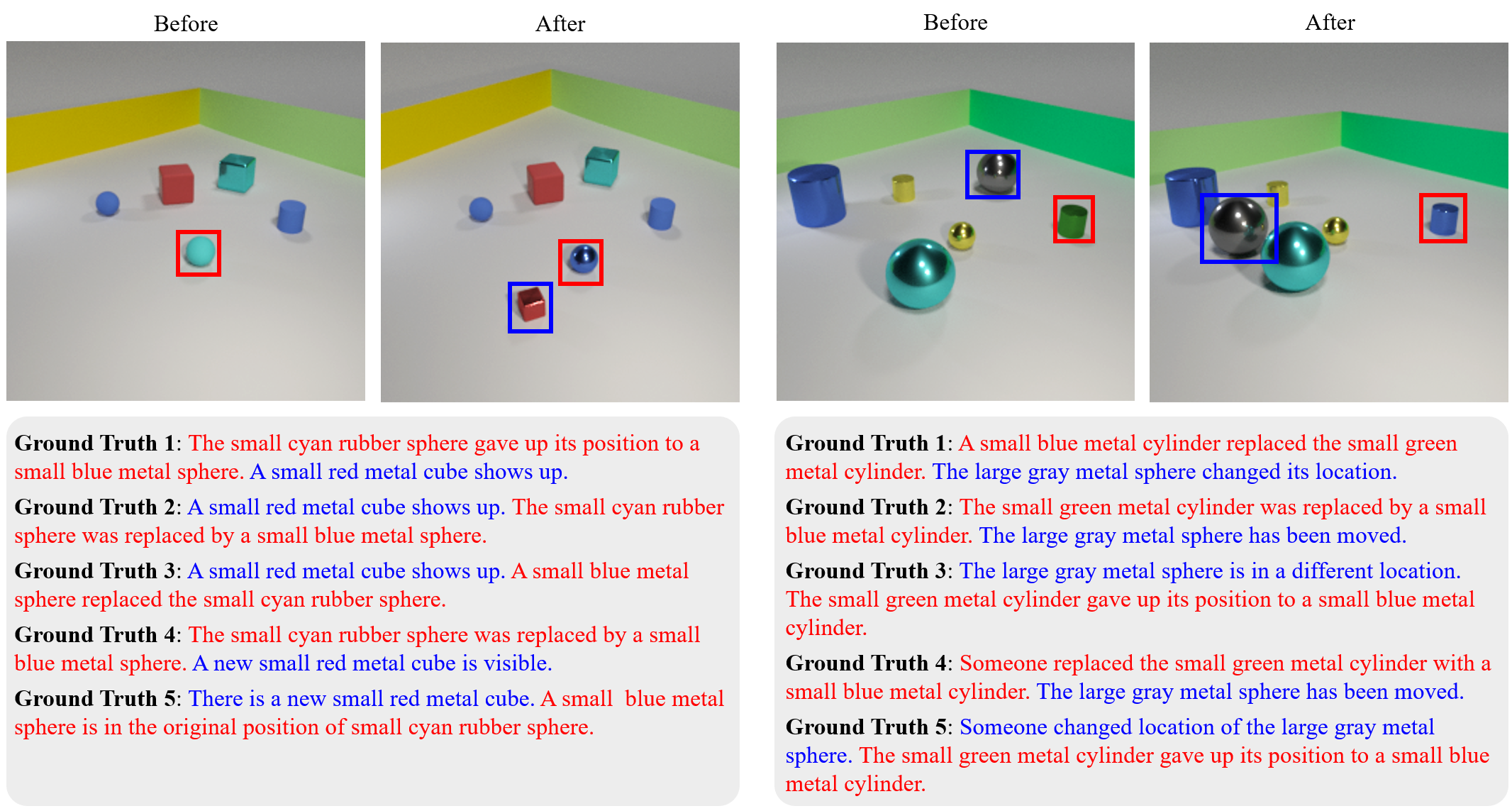}
\caption{Two-change examples from the CLEVR-Multi-Change dataset. The changed objects are highlighted by rectangles with the same color as the associated change captions.}\label{fig:2change}
\end{figure*}

\begin{figure*}
\centering
\includegraphics[width=\linewidth]{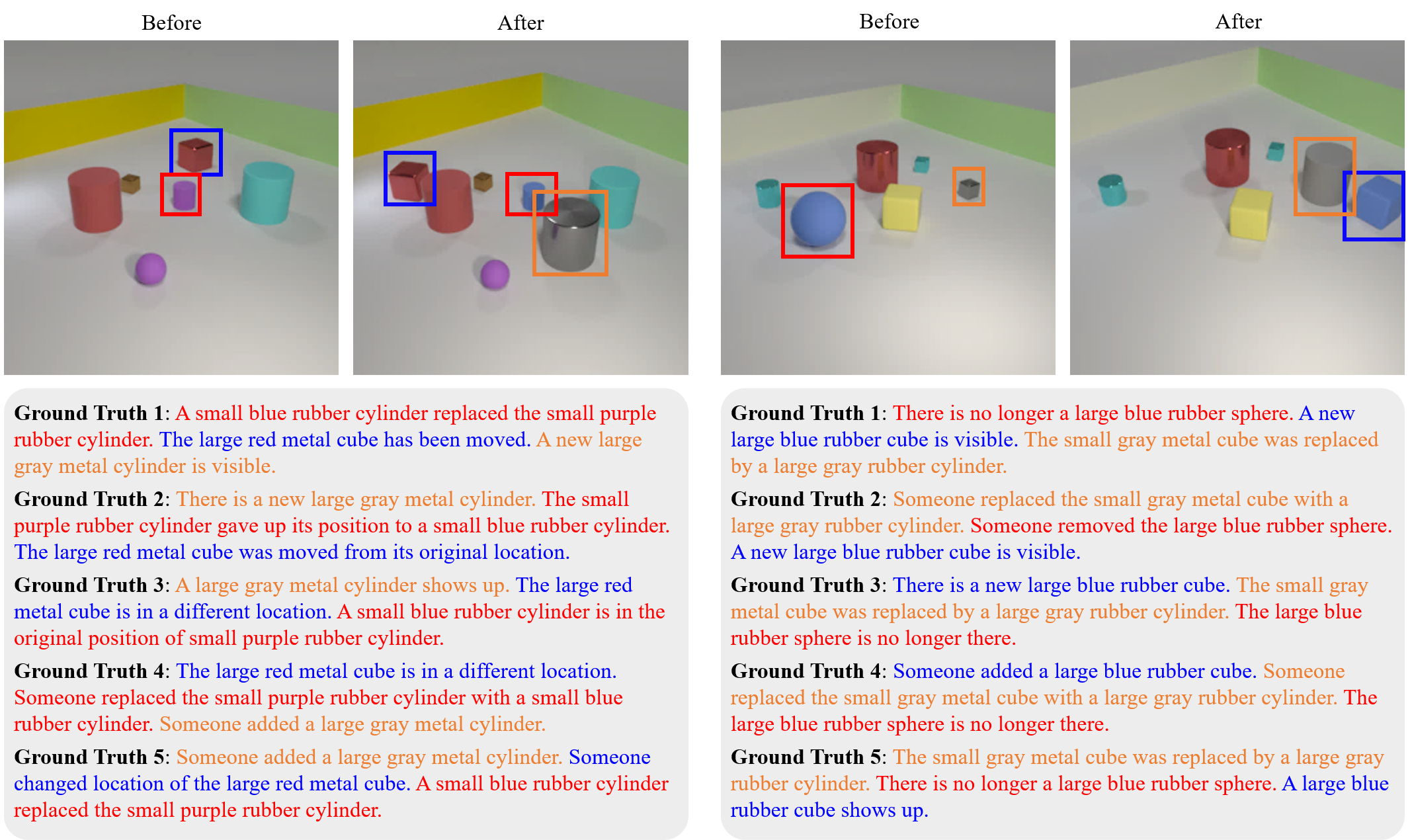}
\caption{Three-change examples from the CLEVR-Multi-Change dataset. The changed objects are highlighted by rectangles with the same color as the associated change captions.}\label{fig:3change}
\end{figure*}

\begin{figure*}
\centering
\includegraphics[width=\linewidth]{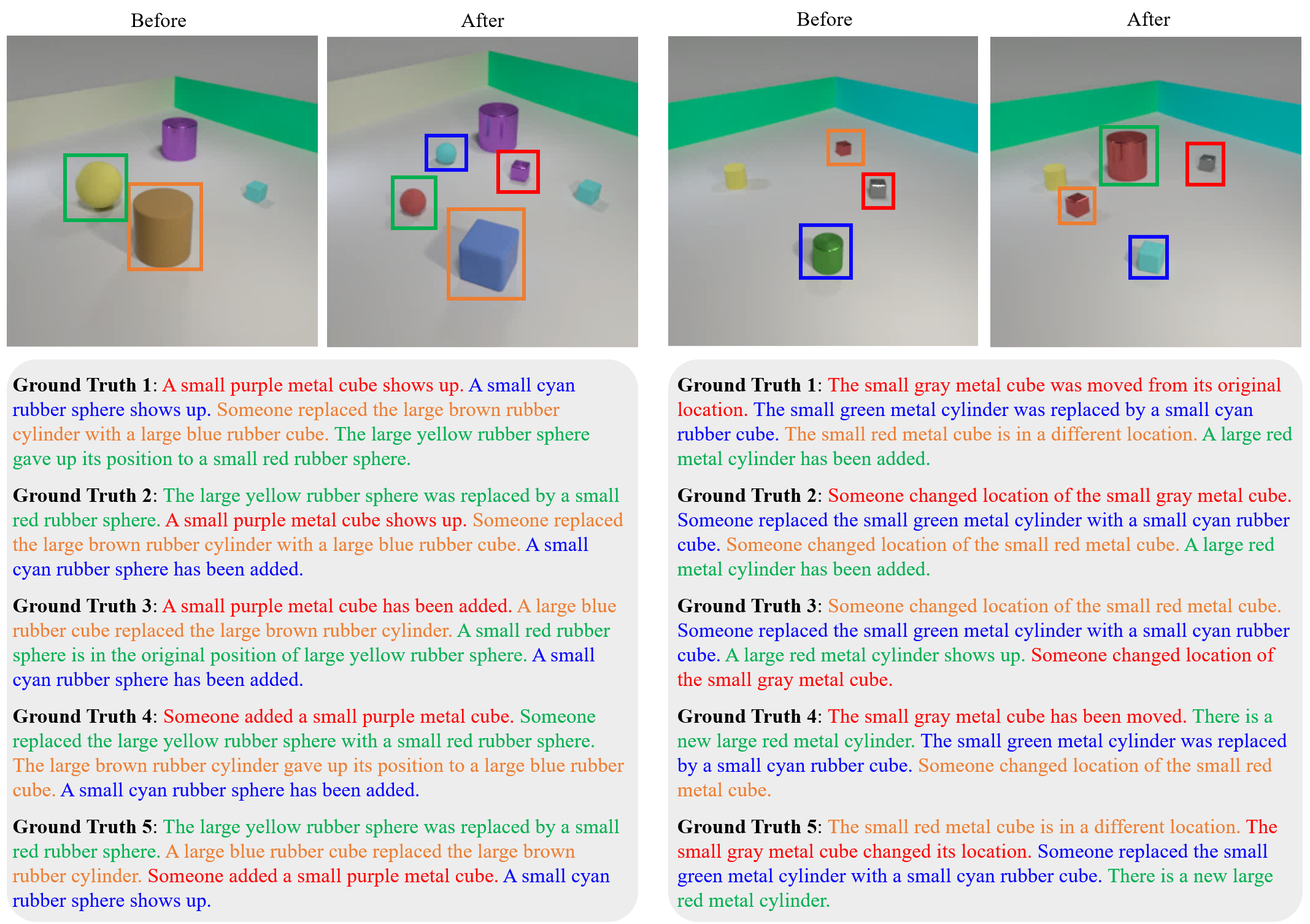}
\caption{Four-change examples from the CLEVR-Multi-Change dataset. The changed objects are highlighted by rectangles with the same color as the associated change captions.}
\label{fig:4change}
\end{figure*}

\textbf{Dataset Examples.} We show additional dataset examples in Figure~\ref{fig:1change} (one-change examples), Figure~\ref{fig:2change} (two-change examples), Figure~\ref{fig:3change} (three-change examples), and Figure~\ref{fig:4change} (four-change examples).

\section{Additional Experimental Results on CLEVR-Multi-Change Dataset}

\begin{figure*}
    \centering
    \includegraphics[width=\linewidth]{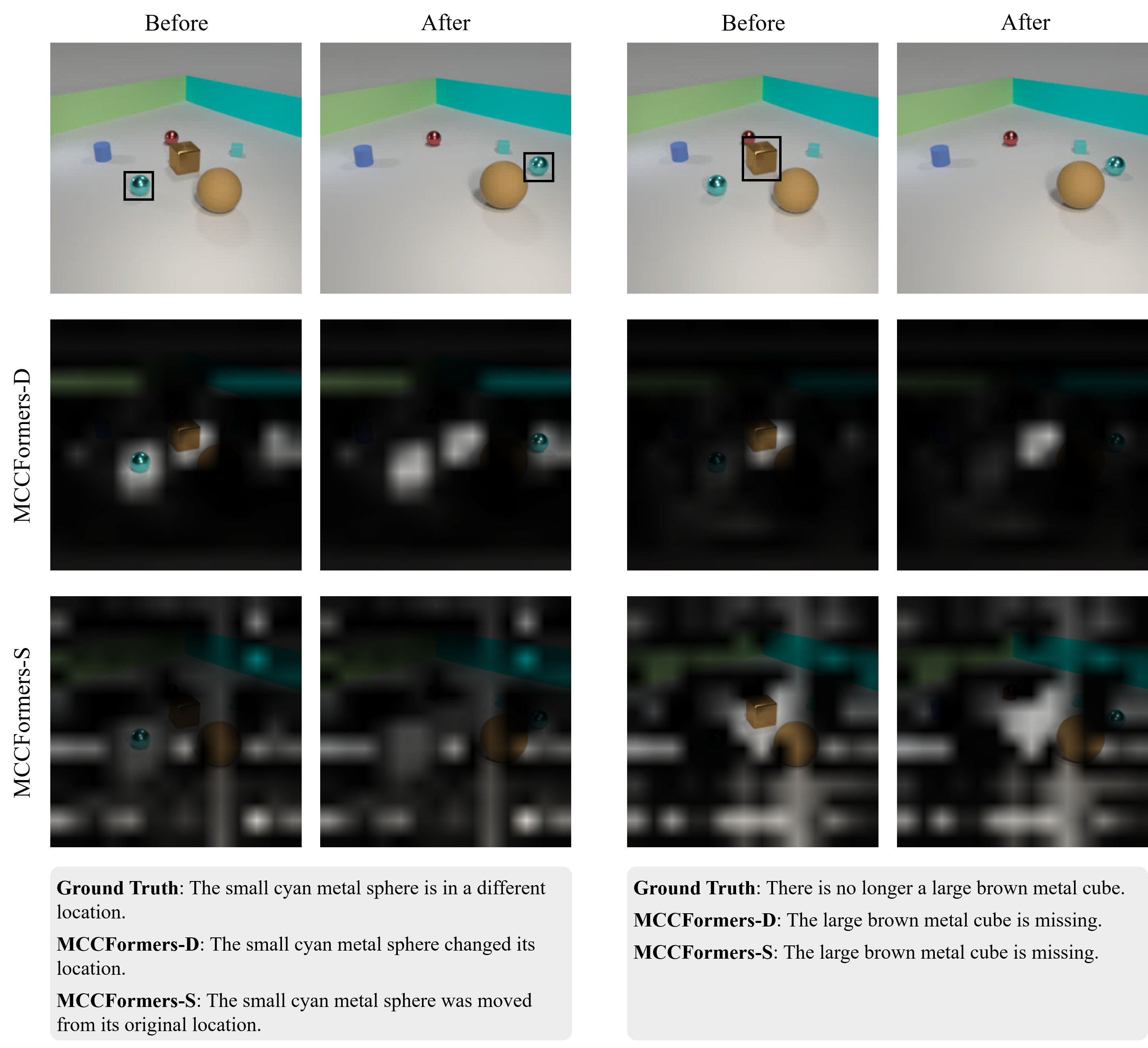}
    \caption{Visualization of an example from the CLEVR-Multi-Change dataset. We highlighted changed regions in black rectangles.}
    \label{fig:result1}
\end{figure*}

\begin{figure*}
    \centering
    \includegraphics[width=\linewidth]{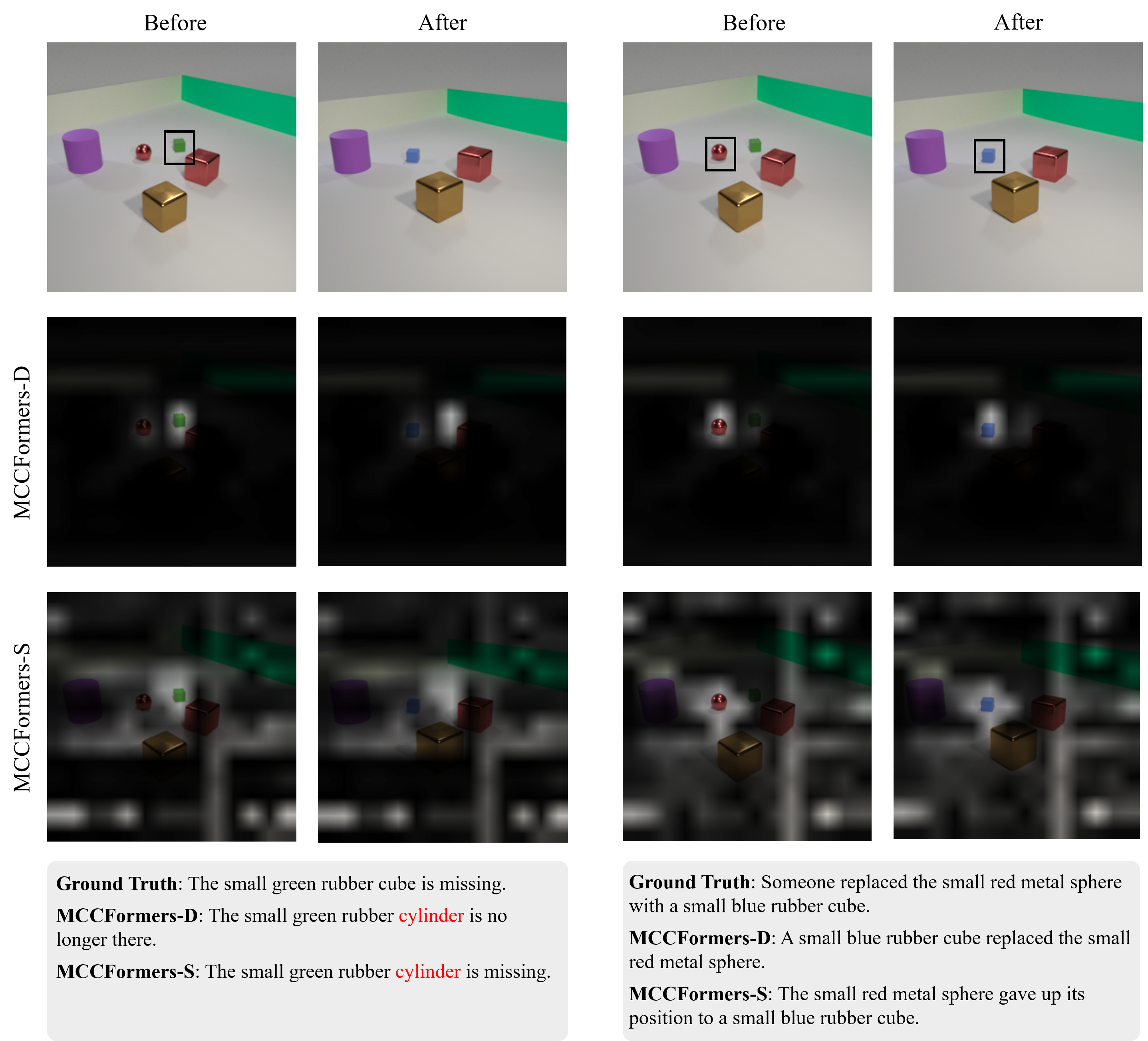}
    \caption{Visualization of an example from the CLEVR-Multi-Change dataset. Incorrect captions are in \cR{red} font. We highlighted changed regions in black rectangles.}
    \label{fig:result2}
\end{figure*}

\begin{figure*}
    \centering
    \includegraphics[width=\linewidth]{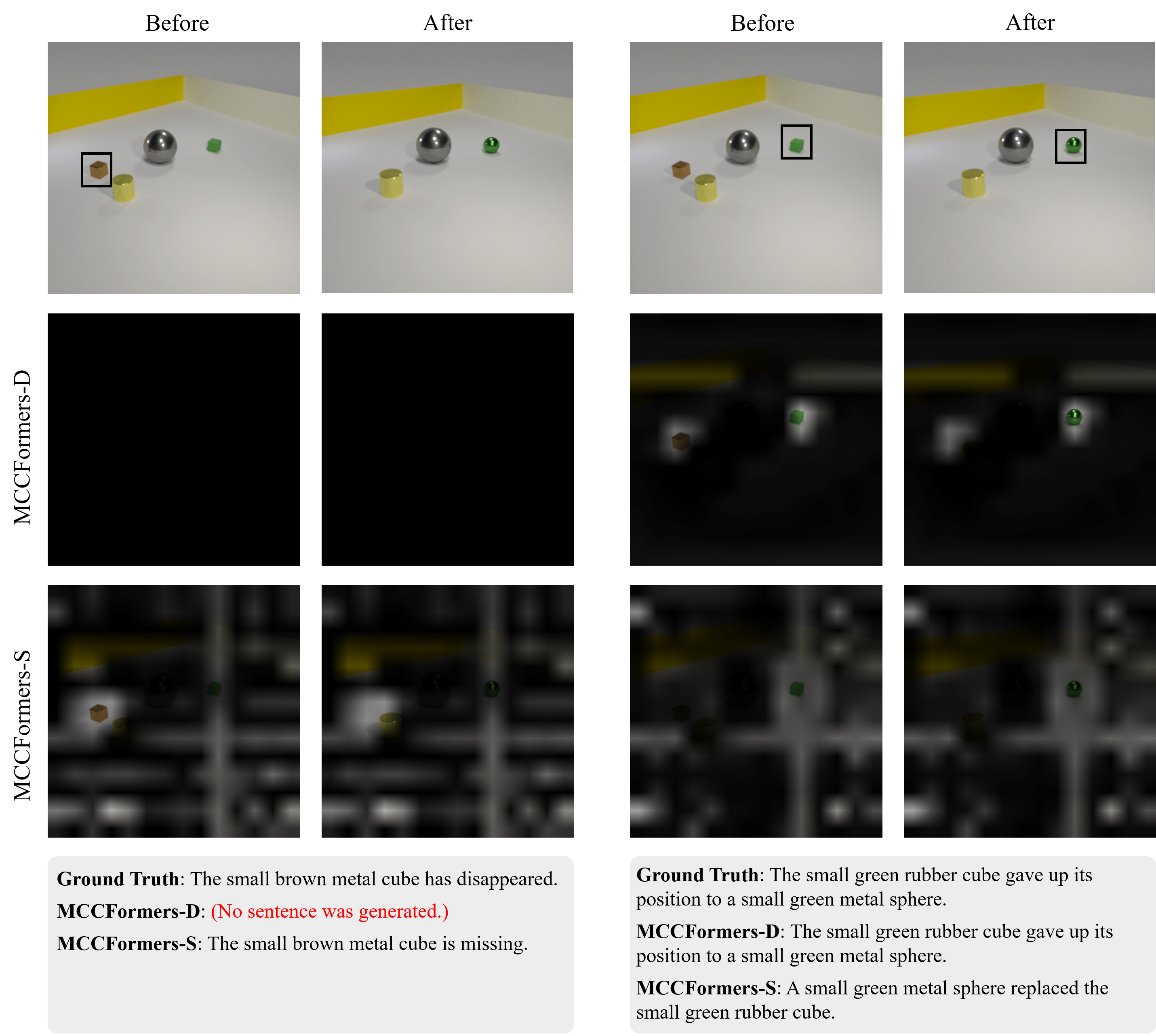}
    \caption{Visualization of an example from the CLEVR-Multi-Change dataset. Incorrect captions are in \cR{red} font. We highlighted changed regions in black rectangles.}
    \label{fig:result3}
\end{figure*}

\textbf{Additional Visualization of Examples.} We show three examples with two changes on the CLEVR-Multi-Change dataset in Figure~\ref{fig:result1}, Figure~\ref{fig:result2}, and Figure~\ref{fig:result3}. For the first two examples (Figure~\ref{fig:result1} and Figure~\ref{fig:result2}), both two MCCFormers correctly generated two related sentences, while for the second example, both two MCCFormers generated a sentence with incorrect object shapes. For the third example (Figure~\ref{fig:result3}), MCCFormers-D only generated one sentence, while the attention maps show that the model captured two change regions. 

Overall, MCCFormers-D obtained attention maps that attend to related change regions while the MCCFormers-S tends to attend to related change regions as well as unrelated regions.

\textbf{Alations of Network Design of MCCFormers (Layers and Heads).} The overall BLEU-4 scores of MCCFormers-D and MCCFormers-S with different layers and heads are shown in Table~\ref{table9}. We found that models with two layers and four heads perform relatively well for both two methods among different network designs. Therefore, we used MCCFormers-D and MCCFormers-S with two layers and four heads in experiments described in the main paper.

\newpage

\newpage

\newpage

\newpage

\newpage

\newpage

\clearpage

{\small
\bibliographystyle{unsrt}
\bibliography{egbib}
}

\end{document}